\documentclass[a4paper, 10pt, conference, table]{ieeeconf}      % Use this line for a4
\usepackage{FG2018}
\usepackage[utf8]{inputenc}
\usepackage{subcaption}
\usepackage{booktabs}
\usepackage{colortbl}
\usepackage{graphicx}
\usepackage[ruled]{algorithm2e}
\usepackage{amsmath}
\usepackage{amssymb}
\usepackage{xcolor}
\usepackage{collcell}
\usepackage{hhline}
\usepackage{pgf}
\usepackage{multirow}

\usepackage[normalem]{ulem} % For striking out text (with sout command)
\makeatletter
\let\NAT@parse\undefined
\makeatother
\usepackage{hyperref}

\def\colorModel{hsb} %You can use rgb or hsb

\newcommand\ColCell[1]{
  \pgfmathparse{#1<50?1:0}  %Threshold for changing the font color into the cells
    \ifnum\pgfmathresult=0\relax\color{white}\fi
  \pgfmathsetmacro\compA{0}      %Component R or H
  \pgfmathsetmacro\compB{#1/100} %Component G or S
  \pgfmathsetmacro\compC{1}      %Component B or B
  \edef\x{\noexpand\centering\noexpand\cellcolor[\colorModel]{\compA,\compB,\compC}}\x #1
  }
\newcolumntype{E}{>{\collectcell\ColCell}m{0.4cm}<{\endcollectcell}}  %Cell width

\definecolor{cwblue1}{rgb}{0.27,0.427,0.623}
\definecolor{cwblue2}{rgb}{0.286,0.454,0.658}
\definecolor{cwblue3}{rgb}{0.733,0.811,0.905}

\newcommand{\bb}{\mbox{\boldmath$b$}}

\newcommand{\bo}{\mbox{\boldmath$o$}}
\newcommand{\bp}{\mbox{\boldmath$p$}}
\newcommand{\bq}{\mbox{\boldmath$q$}}
\newcommand{\bu}{\mbox{\boldmath$u$}}

\def\DBLong{Itekube-7 Touch Gestures Dataset}

\FGfinalcopy % *** Uncomment this line for the final submission

\IEEEoverridecommandlockouts                              % This command is only
\def\FGPaperID{13} % *** Enter the FG 2018 Paper ID here

\title{\LARGE \bf
Learning to recognize touch gestures: recurrent vs. convolutional features and dynamic sampling
}

\begin{document}

\ifFGfinal
% \IEEEoverridecommandlockouts\pubid{\makebox[\columnwidth]{978-1-5386-2335-0/18/\$31.00~\copyright{}2018 IEEE \hfill}
% \hspace{\columnsep}\makebox[\columnwidth]{ }}
\thispagestyle{plain}
\pagestyle{plain}
\author{\parbox{16cm}{\centering
    {\large Quentin Debard$^{1,4}$, Christian Wolf$^{2,1}$, St\'e{}phane Canu$^3$ and Julien Arn\'e{}$^4$}\\
    {\normalsize
    $^1$ Univ de Lyon, INSA-Lyon, CNRS, LIRIS, F-69621, France\\
    $^2$ Univ de Lyon, INSA-Lyon, INRIA, CITI, F-69621, France\\
    $^3$ Normandie Univ, INSA Rouen, UNIROUEN, UNIHAVRE,  LITIS, France\\
    $^4$ Itekube, France}}}
\else
\author{\parbox{16cm}{\centering
    }Anonymous FG 2018 submission \\ Paper ID \FGPaperID \\}
\pagestyle{plain}
\fi
\maketitle

%%%%%%%%%%%%%%%%%%%%%%%%%%%%%%%%%%%%%%%%%%%%%%%%%%%%%%%%%%%%%%%%%%%%%%%%%%%%%%%%
\begin{abstract}
%We propose a fully automatic method for learning gestures on big touch tables used by potentially multiple people. The goal is to learn general models which are capable of adapting to different user styles and hardware variations (table sizes, sampling frequencies and regularities etc.).
We propose a fully automatic method for learning gestures on big touch devices in a potentially multi-user context. The goal is to learn general models capable of adapting to different gestures, user styles and hardware variations (e.g. device sizes, sampling frequencies and regularities).
%
%Based on deep neural networks, our method features a novel dynamic sampling and temporal normalization component, which transforms variable length gestures into fixed length representations and which preserves finger/surface contact transitions, i.e. the topology of the signal. The sequential representation is then dealt with a convolutional model, which, in contrast to recurrent neural networks, is able to extract a hierarchical representation.
Based on deep neural networks, our method features a novel dynamic sampling and temporal normalization component, transforming variable length gestures into fixed length representations while preserving finger/surface contact transitions, that is, the topology of the signal. This sequential representation is then processed with a convolutional model capable, unlike recurrent networks, of learning hierarchical representations with different levels of abstraction.
%
%As another contribution, we introduce the Itekube-7 touch gestures dataset, which is, up to our knowledge, the first publicly available multi-touch gesture dataset for interaction of its size.

To demonstrate the interest of the proposed method, we introduce a new touch gestures dataset with 6591 gestures performed by 27 people, which is, up to our knowledge, the first of its kind: a publicly available multi-touch gesture dataset for interaction.

We also tested our method on a standard dataset of symbolic touch gesture recognition, the MMG dataset, outperforming the state of the art and reporting close to perfect performance.
\end{abstract}

%%%%%%%%%%%%%%%%%%%%%%%%%%%%%%%%%%%%%%%%%%%%%%%%%%%%%%%%%%%%%%%%%%%%%%%%%%%%%%%%
\section{INTRODUCTION}
\label{sec:introduction}

\noindent
Touch screen technology has been widely integrated into many different devices for
about a decade, becoming a major interface with different use cases
ranging from smartphones to big touch tables.
Starting with simple interactions, such as taps or single touch gestures, we
are now using these interfaces to perform more and more complex actions,
involving multiple touches and/or multiple users. If simple interactions do not
require complicated engineering to perform well, advanced manipulations such as
navigating through a 3D modelisation or designing a document in parallel
with different users still craves for easier and better interactions.

As of today, different methods and frameworks for touch gesture recognition were
developed (see for instance \cite{Kin2012}, \cite{Scholliers2010} and \cite{Cirelli2014}
for reviews). These methods define
a specific model for the class, and it is up to the user to execute the correct
protocol. Our approach in this paper is to let users define gestures from a
simplified protocol. The main motivation is to remove burden from the user
and put it onto the system, which shall learn how users perform gestures.
The idea is not new and was first explored in 1991 by Rubine \cite{Rubine1991},
using Linear Discriminant Analysis (LDA) on 13 handcrafted features. Although other
methods discussed in section \ref{sec:related} have built on this idea, our
goal is to generalize it even further by learning deep hierarchical
representations automatically from training data.

\begin{figure}[t] \centering
   \includegraphics[width=7cm]{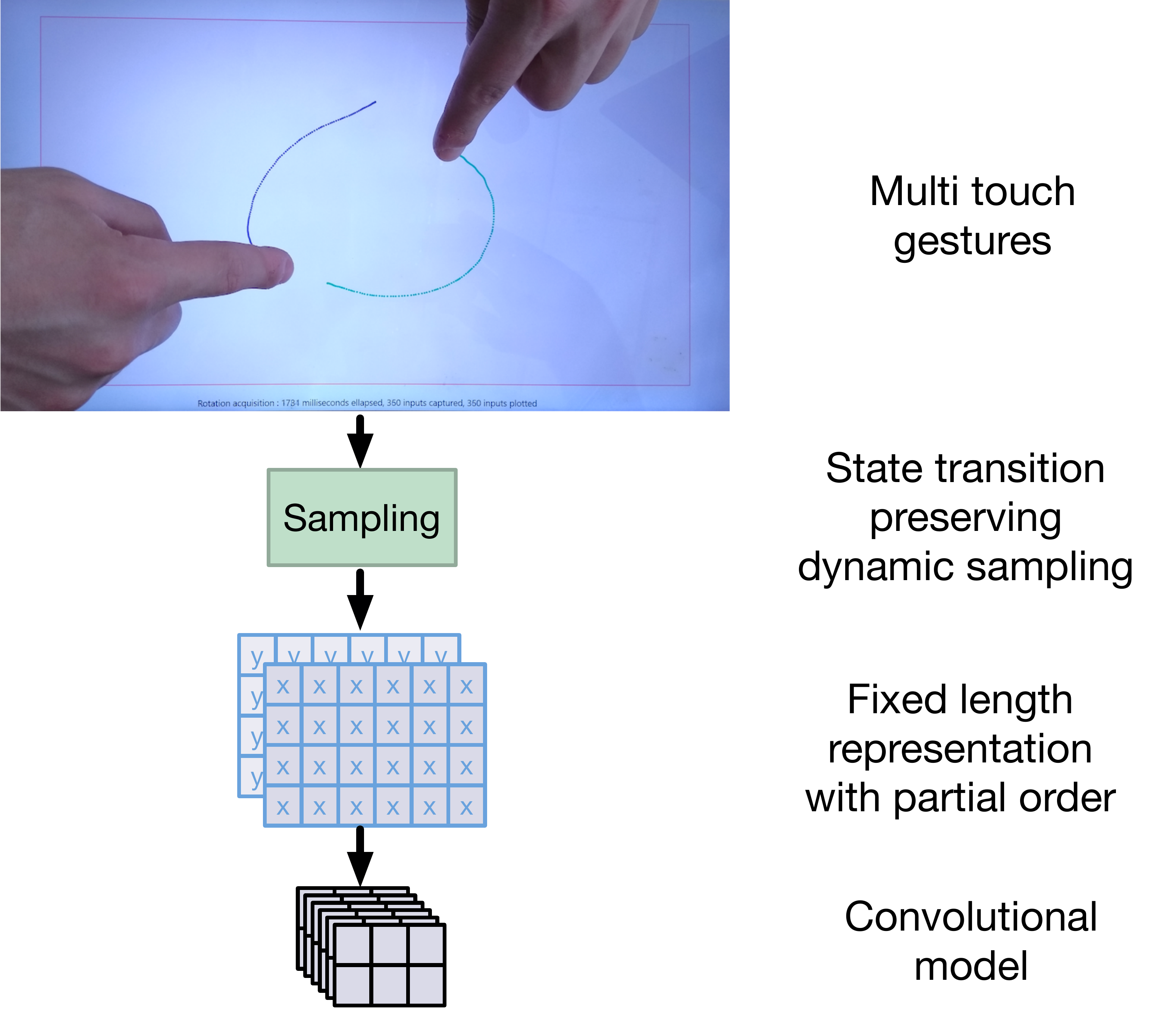}
   \caption{\label{fig:teaser}Overview of the method: multi-touch gestures are dynamically sampled with a transition preserving transform followed by convolutional or 2D-recurrent representation learning.}
\end{figure}
%In order to be adaptative, our method needs to be robust to varying sampling frequencies, screen size, user preferences, and still be able to categorize closely related classes. With the recent advances in sequential learning, deep neural networks are logically to be considered as a potential solution for this problem.
To achieve these goals, our method must capture the particularities of each
gesture class while at the same time being robust with respect to variations
in sampling frequencies, screen size, user preferences.
A further constraint is the computational complexity of the prediction model,
since decoding our model must be possible in real-time applications.

%Another objective is to learn feature representations automatically from data in order to avoid handcrafting features. We propose a

%With recent advances in sequential learning, deep nerve networks must be seen as a potential solution to this problem.
%
%Also, an important limitation when using deep learning models is available data, and
%there is not much public data available to train such a deep architecture.

To address these issues, we propose a new method and provide several contributions:
\begin{itemize}
  \item We address the problem of classifying sequential data characterized by variable amounts of data per time instant. We propose a convolutional model for this problem, and show that it is able to outperform classically used recurrent neural networks on this task. In contrast to recurrent models, our convolutional model is able to create hierarchical representations encoding different levels of abstraction.

  \item We propose a novel input sampling method which drastically reduces the length of the input sequences while at the same time perserving important sharp transitions in the data.

  \item We propose a new dataset of multi-touch sequential gestures, which is, up to our knowledge, the first of its kind. Existing datasets are restricted to symbolic gestures. The dataset will be made publicly available on acceptance of the paper.

  \item We also compare the method against the state of the art on a standard dataset in touch gesture recognition.
\end{itemize}

%%%%%%%%%%%%%%%%%%%%%%%%%%%%%%%%%%%%%%%%%%%%%%%%%%%%%%%%%%%%%%%%%%%%%%%%%%%%%%%%
\section{Related work}
\label{sec:related}

\noindent
Automatic human gesture recognition is an extremly prolific field. Using many
different sensors such as RGB cameras, depth sensors, body sensors, in our case
touch surfaces, these gestures are classified and measured through representations:
geometric features, graphs, state machines, sequences, and more recently, learned features.
The classifying and measuring algorithms are also varied, ranging from deterministic
decisions to Support Vector Machines and Deep Neural Networks.

{\textbf{Touch gestures} --- } can be distinguished into two types:
\begin{itemize}
 \item \textbf{symbols}, such as drawings or handwriting. These gestures are spatially
 complex, but their temporal properties are of lesser interest.
 \item \textbf{interactions}, meant to perform an action using the touch surface as an
 interface. These actions require spatial and temporal precision, as the user
 will expect the interaction to be as precise and fast as possible.
\end{itemize}
Touch gestures are traditionally dealt with handcrafted representations. The most
commonly used methods have been developed by system designers, using procedural
event-handling algorithms (see for instance \cite{Wu2003} or \cite{Malik2005}). Different
frameworks such as Gesture Markup Language (GestureML) were proposed in order to formalize
touch gesture interactions. Midas \cite{Scholliers2010} uses a set of logical rules to classify events on
the surface. With the possibility to define custom operators and priority rules,
its gesture definition is extensible to some point, but lacks rotation invariance.
Proton++ \cite{Kin2012} is another framework, based on regular expressions for gesture
recognition: a gesture is seen as a sequence of events. However, it only supports a unique
gesture at a time, and is limited by the rigidity of regular expressions.

As efficient and fast as they can be, these methods arbitrarily declare gesture properties,
making the user adapt to them. The gestures are precisely defined and tend to lack
generalization in a different context; this contradicts our paradigm of minimal user constraint
and maximum generalization.

In contrast to these user-defined gesture frameworks, Rubine in 1991 developed
a more flexible gesture definition \cite{Rubine1991}, using handcrafted geometric features and
LDA for classification. Up to our knowledge, this is the first attempt at
classifying gestures using deep learning.
Gesture Coder~\cite{Lu2012} takes a similar approach as Proton++, as it defines gestures
using state machines, equivalent to regular expressions on ``atomic actions''.
However, these state machines are learnt from user gestures. \cite{Chen2014} uses a graph representation
of gestures, then embeds the graph structure into a feature vector. These feature vectors
are then classified using a Support Vector Machine.

We also recommend \cite{Cirelli2014} as a good survey of the evolution in multi-touch recognition.

{\textbf{Visual / Air gestures} --- } are gestures performed without any touch surface and captured by video cameras or depth sensors. We mention these methods here, since a part of the methods described in the literature can be adapted to touch gesture recognition, in particular if the input video is first transformed into a structured representation through articulated pose estimation (a skeleton). We will only briefly mention methods using handcrafted representations, which normalize a skeleton into a view- and person-invariant description, followed by machine learning \cite{2013movingpose}, as well as methods based on deep neural networks working on pose alone \cite{Liu2016} or pose fused with raw video \cite{NeverovaWolfTaylorNeboutPAMI2016,LiNeverovaWolfTaylor2017}.

{\textbf{Sequential models} --- } are the main methodology for the gesture recognition, gesture data
being of sequential nature. One of the first successful statistical models used
are Hidden Markov Models (HMMs) \cite{Rabiner1989}, which are generative probabilistic graphical models with a linear chain structure. The hidden state of these models is stochastic, therefore, in the most frequent variants, training and decoding requires to solve combinatorial problems. %These models are still actively used, mainly in biology and natural language processing, but evolved because
%of their limitations:
Conditional Random Fields (CRFs) \cite{Lafferty2001} are discriminative counterparts of HMMs. The modeled distribution is conditioned on the observations, which allows the model to concentrate its capacity on the discrimination problem itself.

Recurrent Neural Networks (RNNs) are the connexionist variant of sequential models introduced in the 80s. Their hidden state is rich and componential and not stochastic, which allows to decode the model in a forward pass as a computation in a direct acyclic graph. An important variant of RNNs are Long Short-Term Mermory networks (LSTMs) \cite{Hochreiter97}, which models additional long term transitions through gates in the model, and their simplier variations GRU \cite{cho-al-emnlp14}.
Clock-work RNNs, introduced in \cite{KoutnikGGS14}, introduce sampling and update frequencies into recurrent networks, allowing for a hierarchical decomposition of the signal. Dense CWRNN adapt this representation to shift invariant models \cite{NeverovaArxiv2016googleIEEEAccess}.

%{\textbf{Convolutional features for sequences} --- } ...
%For activity recognition \cite{BaradelArxiv2017}, for machine translation(publi de FB) \cite{GehringAGYD17}

%%%
\begin{figure*}[t] \centering
   \includegraphics[width=50em]{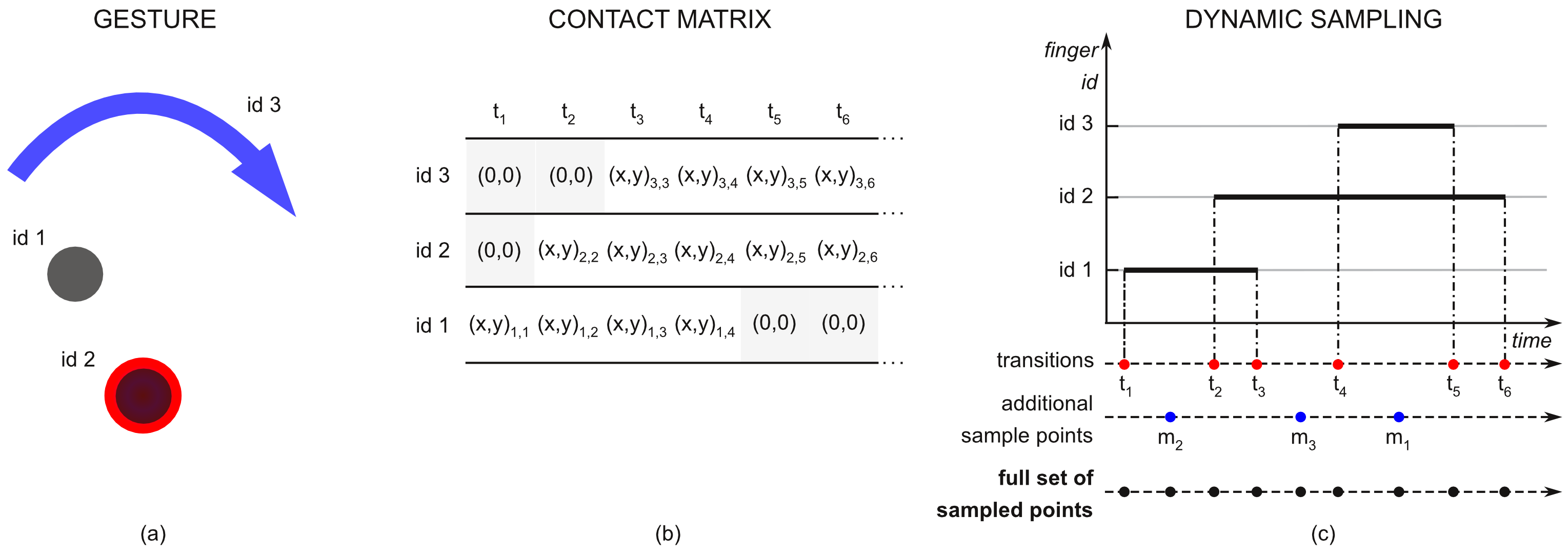}
   \caption{\label{fig:process}From raw data to gesture representations: (a) a graphical representation of a gesture, where red presses are held throughout the gesture; grey touches are taps: quick, almost immobile contacts with
   the surface; blue touches are slides. (b) the device delivers sequences of 2D coordinates $(x,y)$ over time and for tracked finger IDs; (c) Our sampling procedure (section \ref{sec:sampling}) selects samples time @instants $t$ at finger state transitions (in red) and additional sample points (blue).}
\end{figure*}
%%%

%%%%%%%%%%%%%%%%%%%%%%%%%%%%%%%%%%%%%%%%%%%%%%%%%%%%%%%%%%%%%%%%%%%%%%%%%%%%%%%%
\section{Recurrent vs. convolutional models}
\label{sec:method}

\noindent
Our problem can be cast as a sequential learning problem with input dimensions varying over time.
Each input gesture is a sequence $(\bu^n_{t,i})_{t=1,T_n}$ of length $T_n$
where $n$ is the index of the touch gesture,
$i$ the finger ID provided by the touch device and $\bu^n_{t,i} = (x^n_{t,i} \,,\, y^n_{t,i} )^T$ are the spatial 2D coordinates of finger $i$ on the touch screen. An example of such a sequence is illustrated schematically in Figure \ref{fig:process}a.
Note that a variable amount of fingers may touch the surface. Therefore, finger ID indexes $i \in \{1\dots I_n \}$, $I_n$ being the number of fingers involved in gesture $n$.
Finger IDs are from an unordered set and provided directly by the touch hardware. We suppose that finger tracking allows finger IDs to be identical for the same finger as long as the finger touches the screen; however, removing a finger and putting it on the screen again will not generally preserve its ID. A similar approach was taken in \cite{Rubine1991}.
In the following, gesture indices $n$ can be omitted for clarity, unless necessary for comprehension.

%Our problem can be cast as a sequential learning problem with input dimensions varying over time. In particular, our input gestures $\bu^n_{t,i} = [ x^n_{t,i} \ y^n_{t,i} ]^T$ are spatial 2D coordinates of finger positions on the touch screen indexed over gesture index $n$, time $t$ and over finger Id $i$. Finger Ids are from an unordered set and provided directly by the touch hardware. We suppose that finger tracking allows finger Ids to be identical for the same finger as long as the finger touches the sceen; however, removing a finger and putting it on the screen again will not generally preserve its Id. A similar approach was taken in \cite{Rubine1991}.

%Each gesture is a squence of length $T_n$. At each instant $t$, a variable amount of fingers may touch the surface. Therefore, finger Id indexes $i \in \{1\dots I_n \}$. In the following, gesture indices $n$ can be omitted for clairy, unless necessary for comprehension.

%\subsection{Convolutional features for sequences}

%\noindent
We address the problem of joint learning of a prediction model for classification together with a feature representation from training data. The main difficulty we face is the fact that the data is temporal with a variable number of data points (fingers) at each time step, which makes it difficult to align off-the-shelf sequential models like RNNs and their variants directly on the input feature dimension of the data.

%\textbf{Learning representations of unordered data ---}
One strategy is to train a model, which integrates data points $\bu_{t,i}$ for a single time instant $t$ into a fixed length representation through a learned mapping $f_t=\phi(\bu_{t,1:I_t},\theta_\phi)$ parameterized by $\theta_\phi$ (gesture index $n$ has been omitted). Again, the index $1{:}I_t$ indicates that the number of inputs, which are of variable length depending on $t$. This makes it difficult to learn this mapping with classical models, which suppose that the data are embedded in a vector space. In our case, each data point is of fixed length, but the set indexed by $1{:}I_t$ is not.

Handcrafted representations could be designed, to embed this set of samples into a fixed length representation which describes the spatial distribution of the points. In the literature, several representations have been proposed, but we will only mention Shape Context \cite{BelongieMalik2002}. In \cite{BelongieMalik2002}, log polar histograms are computed for a point cloud, which describe the positions of individual points relatively to other points in the cloud.

In our work, we prefer to automatically learn a suitable feature representation from data. One way is to integrate the different samples $\bu_{t,i}$ iteratively over $i$ using a sequential model. Note, that this integration over data points (fingers) is here done for a single time instant $t$.

It is very important to remark here, that a sequential model is trained on data which is \emph{unordered}. In other words, the model is required to ignore the evolution of the data over it's finger ID dimension, since the input data is not ordered in this dimension. The model will need to learn to embed the data into a spatial representation, in a similar spirit as Shape Context histograms. Ensuring invariance to finger order is therefore important, whose learning can be favored with data augmentation techniques, i.e. shuffling finger IDs during training.

\begin{figure*}[t] \centering
\includegraphics[width=18cm]{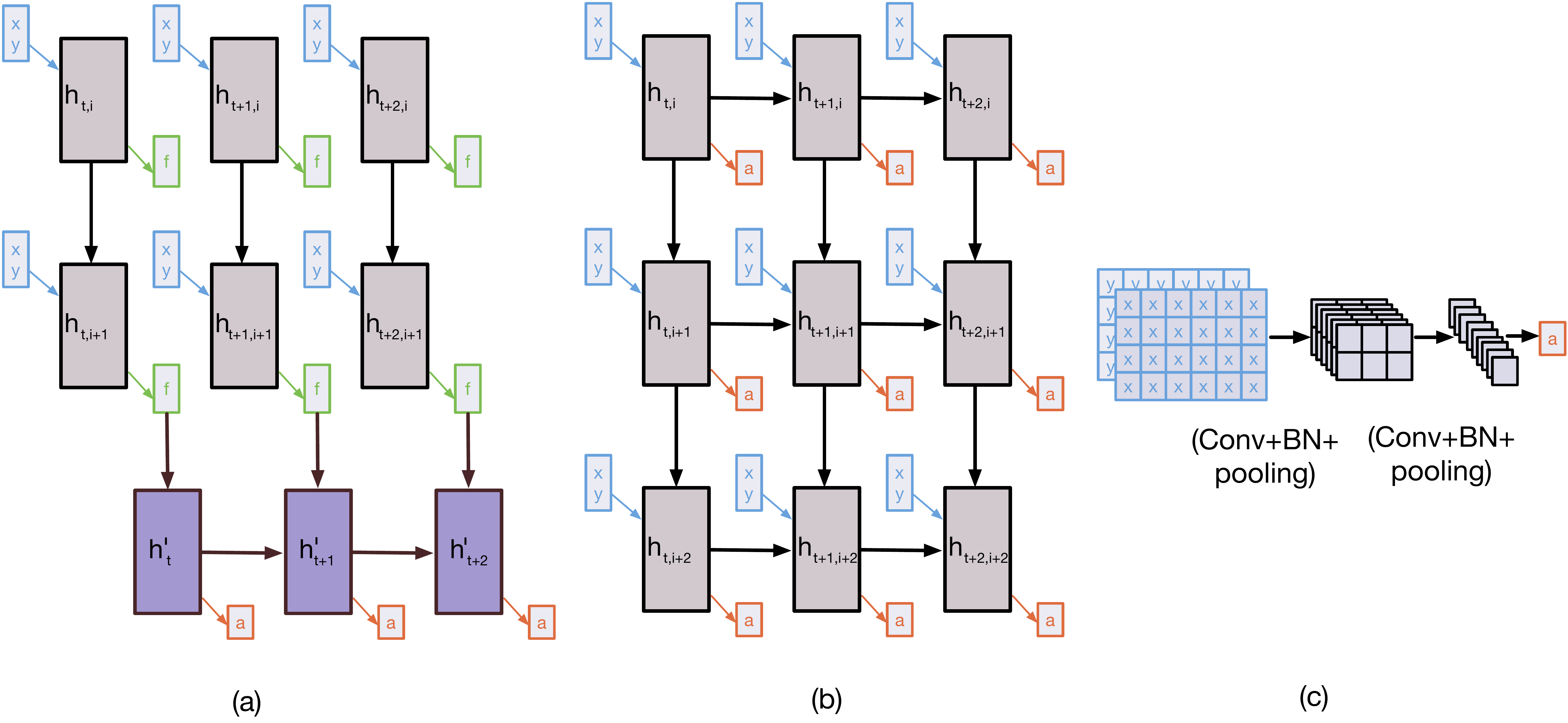}
% \begin{tabular}{ccc}
%     \begin{minipage}{5cm}
%     \includegraphics[width=5cm]{inserthere.jpg}
%     \end{minipage}
%     &
%     \begin{minipage}{5cm}
%     \includegraphics[width=5cm]{inserthere.jpg}
%     \end{minipage}
%     &
%     \begin{minipage}{5cm}
%     \includegraphics[width=5cm]{inserthere.jpg}
%     \end{minipage}
%     \\
% \end{tabular}
\caption{\label{fig:representation}Feature representations for sequences: (a) sequential learning of fixed-length representations; (b) MD-RNNs / MD-LSTMs (gates are not shown); (c) convolutional features for sequential data. Input data is shown in blue, output in red (best viewed in color).}
\end{figure*}

The resulting features $f_t$ can then be integrated temporally using a second sequential model $a{=}\psi(f_{1:T}, \theta_\psi)$, parametrized by $\theta_\psi$, which operates in the time dimension and predicts a gesture class $a$ for the sequence. This model is illustrated in figure \ref{fig:representation}a, which shows the two mappings $\phi(.)$ and $\psi(.)$ as recurrent networks with respective hidden units $h$ and $h'$. Here, features $f_t$ are outputs of $\phi(.)$ --- an alternative choice would have been to use the hidden representation $h$ itself as features input to $\psi(.)$.

The model described above corresponds to the basic requirements of a model responding to the given problem. In the rest of this section, we will provide two deep neural models extending this principle, namely recurrent networks and convolutional neural networks. We will argue the superiority of convolutional features, and in section \ref{sec:experiments} we will confirm these arguments through experiments.

\textbf{A multi-dimensional recurrent model ---}
As mentioned above, touch finger IDs are not ordered; however, fingers are usually tracked over time by the hardware. Finger IDs are therefore consistent over time, at least between finger state transitions. The model described above does not directly describe transitions of individual finger inputs over time. Of course the representation mapping
$\phi(.)$ can theoretically learn a representation $f_t$, which allows the temporal mapping $\psi(.)$ to disentangle individual fingers and track their temporal evolution. In practice, learning this is unnecessarily hard.

A different strategy is to create trainable and direct temporal transitions between successive samples of the same fingers. This requires to handle transitions in two different dimensions: finger ID and time. Using connectionist learning frameworks, one possible strategy is to use multi-dimensional RNNs or LSTMs \cite{Graves2007}\cite{Liu2016}, a straight-forward extension of the 1D original models, or closely related variants like Grid RNNs \cite{Kalchbrenner2016}.
%\footnote{Equivalent models with this topology have also been proposed in the graphical model literature in the form of 2D HMMs and CRFs\cite{XXX-Chris-to-cite}.}.
In these models, the hidden state (and also the hidden memory cell in the case of the LSTM variant of the model) of time $t$ is not only computed from one single predecessor ($t{-}1$) but from several predecessors, one for each dimension. This is illustrated in figure \ref{fig:representation}b for two dimensions (finger ID and time).

This model can handle data of variable length in time and of variable numbers of fingers per time instant. However, when finger state transitions (finger press/release events) occur, partially empty data rows are created, which need to be padded, for instance with zero values.

\textbf{A convolutional model ---} gestures are characterized through their short term behavior as well as their (relatively) long term behavior. A good model should be able to capture the former as well as the latter, i.e. the statistical long range dependencies in the input data. In principle, the recurrent models described above are able to do that. However, since they satisfy the Markov property (the state of the model at time $t$ depends only on the state of the model at time $t{-}1$ and not on the preceding states), all long range dependencies need to be captured by the hidden state.

Convolutional Neural Networks (CNNs), on the other hand, have proven to be efficient in learning hierarchical hidden representations with increasing levels of abstraction in their subsequent layers. Traditionally, these models have been applied to images \cite{Lecun98}, where filters correspond to spatially meaningful operations.

Very recently only, they have been reported to be efficient for learning hierarchical features on sequential data, for instance for human activity recognition from articulated pose (skeletons) \cite{BaradelArxiv2017} or for machine translation \cite{GehringAGYD17}. In these cases, the temporal dependencies of the data are not covered by a unique hidden state, but by a series of feature maps, which capture temporal correlations of the data at an increasing level of abstraction. While the first filters capture short term behavior, the last layers capture long range dependencies.

In this work, we propose a similar strategy for the recognition of touch gestures. The proposed CNN uses 2D spatial filters. The sequential input data is structured into a 3D tensor, with the finger ID as first dimension, time as the second dimension, and input channels as the third dimension ($x$ coordinates are in the first channel and $y$ coordinates in the second channel) --- as illustrated in figure \ref{fig:representation}c. The 2D convolutions therefore capture correlations in time as well as correlations over finger IDs. Of course this dimension is not ordered; here, data augmentation provides the means to guide the model towards invariance in the ordering of the input data.

\textbf{Data preparation ---} in this work we propose convolutional features for touch gesture recognition, and in the experimental section we will compare them to MD recurrent models described above, as well as to the state of the art in touch gesture recognition.

CNNs require fixed length representations\footnote{Fully convolutional architectures withstanding, which are predominant in segmentation applications.}. We therefore propose a novel feature preserving dynamic sampling procedure, which will be described in the next section.

\section{Transition preserving dynamic sampling}
\label{sec:sampling}

\begin{algorithm}[t]
\caption{Dynamic sampling of $K$-points}\label{algo_sampling}
\KwIn{a gesture $U=\{ \bu_{t,i} \}, t\in{[0,T]}$ \\
 $\quad \quad \ \ $ transition indicators $O = \{ \bo_{t}  \}, t\in{[0,T]}$}
\KwOut{Ordered and indexed set of sample points $C$}
\vspace{0.2cm}
\begin{enumerate}
  \item $C = \{0\} \cup \{ t : \bo_t = 1 \}$
\vspace{0.2cm}
  % \item \textbf{For} $t$ \textbf{in} $[1,T]$:\\
  % \hspace{0.5cm} \textbf{if}($\{ \bo_{t} \}=1)$:\\
  % % \hspace{0.5cm} \textbf{if}($((\{ \bu_{i,t} \}=0)$and$(\{ \bu_{i,t-1} \}\neq 0))$
  % or \hspace*{0.95cm}$((\{ \bu_{i,t} \}\neq 0)$and$(\{ \bu_{i,t-1} \}=0))$):\\
  % \hspace*{1.5cm} $C = C \cup \{t\}$
\vspace{0.2cm}
  \item \textbf{While} $|U'| < K$:\\
  \hspace{0.5cm} $k = \underset{x\in{[0,T-1]}}{\arg \max}|C_{x+1}- C_x |$\\
  \hspace{0.5cm} $C = C\cup \left \lfloor{\frac{k}{2}}\right \rfloor$
\end{enumerate}
\end{algorithm}

\noindent
The input gestures are of variable length, of varying numbers of fingers per instant, and eventually of varying speed, which is due to variations in user behavior but also to differences in sampling rates. In practice, we observed samples rates between $5$ ms and $25$ ms. Varying speed can be dealt with easily and traditionally by delegating it to the model and to the training procedure using data augmentation, and/or by using temporal multi-resolution representations.

Varying numbers of fingers is coped with by padding. We define a maximum number of fingers
depending on the application, as shown in Figure \ref{fig:process}, and zero pad the unused entries.

Variable temporal length is a different issue, and is a hard problem for convolutional models. In theory it can be dealt with sequential models, like RNNs and their variants. In practice, effects like vanishing/exploding gradients \cite{Bengio1994} tend to limit these models to relatively short sequences, despite efforts to tackle these limitations \cite{Hochreiter97}\cite{cho-al-emnlp14}.

Existing work tends to perform sampling spatially, as for instance in \cite{Wobbrock2007}:
because the task in these papers is to classify symbolic gestures, temporal features
such as state transitions or velocity are of minimal interest. We call state transition the
moment when at least one finger is added onto or withdrawn from the touch surface.
When the task involves interaction gestures \cite{Chen2014} \cite{Lu2012},
dimension reduction is often done through feature extraction/embedding, not sampling.

We chose to normalize the data through a feature preserving transform which compresses it into a fixed length representation. State transitions, which are key features in touch gesture recognition \cite{Lu2012}, are preserved with this sampling strategy.
For gestures with high temporal content, where spatiality does not alter much the classification (such as a
press tap, see Figure \ref{fig:gestures}), missing one of these transitions will most likely
result in a misclassification of the gesture.
Using a uniform sampling, quick transitions such as a tap can be missed.

The goal is to transform a variable length sequence $( \bu^n_{t,i} ), t=\{1..T^n\}$ into a fixed length representation $( \bu'^n_{t,i} ), t=\{1..K\}$. We perform this by choosing $N$ sampling time instants $t$ which are common over the finger IDs $i$. The set sampling points should satisfy two properties:
\begin{itemize}
\item (i) the points should be spaced as uniformly as possible over the time period;
\item (ii) the sampled signal should preserve finger transitions, i.e. transitions (finger up or finger down) should not be lost by the transform.
\end{itemize}

To formalize this, we introduce a transition indicator variable $o_t$ defined as follows: $o_t{=}1$ if a transition occurs at time $t$ (finger touch down or finger release), and $o_t{=}0$ else. Then, the \emph{inverse} problem, namely creating observed gesture sequences from a set of given sample points, can be modeled as a probabilistic generative model, in particular a Hidden Semi-Markov Model \cite{YuHSMM2010} (HSMM) with explicit state duration modelling. Obtaining samples from the observed sequence then corresponds to decoding the sequence of optimal states.

In this formulation, the indicator variables $o_t$ correspond to the observations of the model, whereas the hidden state variables $S_t$ correspond to the sampling periods. Each state variable can take values in the set of hidden states $\Lambda{=}\{1..K\}$, which correspond to the $K$ target samples. The desired target sampling points correspond to instants $t$ where changes occur in the hidden state $S_t$. The transition function of the model is a classical forward model (upper case letters indicate random variables, lower case letters realizations):
\begin{equation}
\begin{array}{lll}
\bq_{(i,d,j}) & \triangleq & P(S_{[t+1:[}{=}j|S_{[t-d+1:t]}=i) = \\
\\
              & = &
    \left \{
    \begin{array}{ll}
    \frac{1}{S} & \textrm{ if } j{=}i+1 \\
    0 & \textrm{else} \\
    \end{array}
    \right .
    \\
\end{array}
\end{equation}
The duration probability distribution encodes above property (i), which aims sampling points equally spaced with a target period of $\frac{T^n}{K}$:
\begin{equation}
\begin{array}{lll}
\bp_{j,d} & \triangleq & P(S_{[t+1:t+d]}{=}j|S_{[t+1:[}=j) = \\
\\
          & = & \frac{1}{Z}(d-\frac{T^n}{K})^2 \\
\end{array}
\end{equation}
where $Z$ is a normalization constant.

The observation probabilities encode the hard constraints on the transitions (above property (ii)):
\begin{equation}
\begin{array}{lll}
\bb_{j,d}(o_{t+1:r+d}) & \triangleq & P(o_{[t+1:t+d]}|S_{[t+1:t+d]}=j) = \\
\\
                       & = &
    \left \{
    \begin{array}{ll}
    0 & \textrm{if } \sum_{j=t+1}^{r+d} o_j > 1 \\
    \frac{1}{Z'} & \textrm{else} \\
    \end{array}
    \right .
    \\
\end{array}
\end{equation}
where Z' is a normalization constant. In other words, sampling periods spanning over more than 1 transition are forbidden, which makes it impossible to lose state features.

If the number of transitions is lower than the number $K$ of desired sampling points, then the Semi-Markov model parametrized by the above equations\footnote{For space reasons, we omitted the initial conditions which ensure that the optimal sequence begins with state $1$ and terminates with state $T$.} can be decoded optimally using the Viterbi algorithm \cite{YuHSMM2010}. Because the complexity is high, we solve the problem faster and heuristically with a greedy algorithm, which first selects all transition points for sampling, and then iteratively adds additional sampling points in the longest remaining sampling intervals (see algorithm \ref{algo_sampling}).

A drawback of the proposed sampling method is the variations in the sampling rate over gestures: since the
sampling is not uniform, we lose velocity information if we only consider spatial coordinates.
One possibility would be to keep time stamp information additionally to coordinates, making it possible for the model to extract the required local velocity. In practice, experiments showed that the resulting gestures are sufficiently equally sampled and adding time stamps did not improve performance.

% \begin{figure}[t]
%    \includegraphics[width=23em]{./img/dynamic_sampling.png}
%    \caption{\label{dynamic sampling} 9 points dynamic sampling on a 3 fingers gesture.
%    We consider in this graph a binary state for each finger: contact with the surface (Red) or absence
%    of contact (Grey).
%    Red dots are transitions, blue dots are iteratively selected by taking the
%    closest point from the middle of the longest segment between two sampled points. In this case, the sequence
%    $[t_1,m_2,t_2,t_3,m_3,t_4,m_1,t_5,t_6]$ is the sampled gesture.}
% \end{figure}

%%%%%%%%%%%%%%%%%%%%%%%%%%%%%%%%%%%%%%%%%%%%%%%%%%%%%%%%%%%%%%%%%%%%%%%%%%%%%%%%
\section{The \DBLong}
\label{sec:dataset}

\begin{figure}[t]
   \includegraphics[width=24.5em]{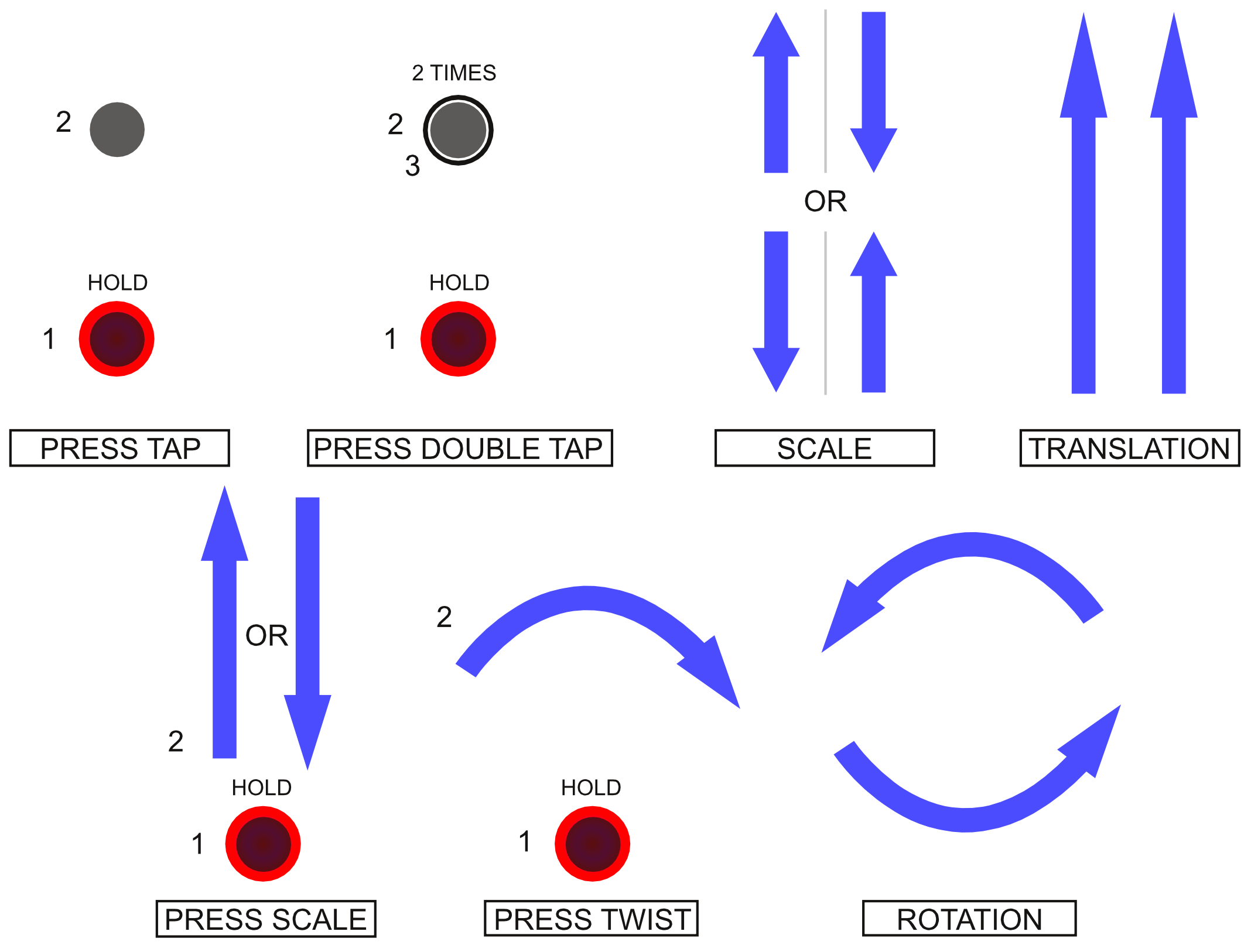}
   \caption{\label{fig:gestures} The classes of the proposed novel multi-touch gestures dataset. Red presses are held
   throughout the gesture. Grey touches are taps: quick, almost immobile contacts with
   the surface. Blue touches are slides.}
\end{figure}

\noindent
We introduce a new touch gesture dataset containing 6591 gestures
of 7 different interaction classes (illustrated in Figure \ref{fig:gestures})
performed by a total of 27 different people.
These persons are from different professional backgrounds and
aged from 12 to 62. The dataset is available at \url{http://itekube7.itekube.com}.
%Each gesture is saved as an xml file with meta information
% and the data.

Samples correspond to finger contacts on the touch table which are
sampled by the touch screen hardware. A sample contains the
finger ID (provided by the hardware), x and y coordinates and a timestamp.
A finger is tracked as long as it
stays in contact with the surface. Coordinates are normalized
with respect to the screen size, from 0 (top-left) to 1(bottom-right).

The gestures descriptions provided to the users were deliberately minimal,
in order to grasp as many user variations as possible.
The gestures can be executed anywhere on the screen,
with any orientation, scale or velocity. Users were asked to perform the gesture naturally, we did not insist on a very strict definition of the finger state transitions.
It means the user knows the different classes, and performs each one as he wants as
long as we can distinguish the different classes. In consequence, some classes can
be defined by different transition sequences; for example, on press tap, some users
lift the press finger first, whereas others lift the tap one first. Some classes are highly correlated:
press tap and press double tap are only distinguishable from their transitions,
press scale and scale differs from one trajectory. From our experiments, these
two classes were usually the hardest to separate (see Table \ref{conf_mat}).

%%%%%%%%%%%%%%%%%%%%%%%%%%%%%%%%%%%%%%%%%%%%%%%%%%%%%%%%%%%%%%%%%%%%%%%%%%%%%%%%

\section{Experimental Results}
\label{sec:experiments}

\noindent
% \begin{description}
We tested our method on 2 different datasets: the \DBLong \hspace{2pt} introduced in section \ref{sec:dataset} and
the Mixed Multistroke Gestures (MMG) dataset \cite{Anthony2012}. The latter dataset contains 9600 gestures from 20
participants in 16 classes. All classes are symbol gestures: the sequence order has no importance
for the classification, we classify the symbol drawn by all the trajectories. Each class is performed
10 times by a participant at three different speeds, resulting in 30 occurrences per class per participant.

% \item[The MatchUp dataset \cite{rekik:hal}] \hspace{28mm} --- 5,155 gestures from 16 participants
% in 22 classes. Again, all the classes are symbol gestures: we could not find a public dataset
% of interaction gestures. Each class can be performed differently: the participant
% can change the number of strokes and or the number of fingers used simultaneously to make the gesture.
% \end{description}
\textbf{Experimental setup} ---
Because of hardware variations, we want our model to be
invariant to finger IDs: we should avoid any correlation between a gesture and the
finger ID provided by the device, which results in a finger order in the input tensor.
To address this problem, we
perform data augmentation on the ID permutations: each gesture of the training
set is augmented by permutating its lines (corresponding to a single finger trajectory). For
our multi-touch dataset, we set the maximum number of fingers to 3 and thus keep all 6 permutations
of each gesture.

The coordinates $(x,y)$ of each datapoint have been normalized
on their respective axis between $0$ and $1$, for any device. This
means that depending on the screen size, gestures will be relatively larger,
smaller or even distorted if the ratio is different.
Normalization proved to be inferior to data augmentation on this problem;
we therefore artificially increase the scale variation of the dataset.
Each gesture permutation is rescaled two times between 0.5 and 1.5.
This brings the size of the augmented training set to 12 times the original one.

Because of the freedom given to the subjects to perform gestures, data augmentation on
gesture orientation was not necessary.

\textbf{Architectures and implementation details} --- We implemented the neural model in
Tensorflow \cite{tensorflow2015}.
% The database is converted
% into Protocol Buffer format (Google's language for serializing structured data) to
% optimize the training process.
All hyper-parameters have been optimized over the validation set, the test set has not been used for this.
In particular, the number of sampled points set to $K{=}10$ was the best trade-off between information maximization and redundancy minimization for our problem.

\begin{itemize}
\item For readability purposes, we refer to convolutional layers with x feature maps as CONVx layers, and
max pooling layers as POOL. The convolutional model has the following architecture:
a CONV128 layer, a $1{\times}2$ POOL layer (max pooling only on the time dimension), again a CONV128 layer and a $1{\times}2$ POOL layer,
a CONV256 layer and a fully connected layer providing the prediction score for each class.
Activation functions are ReLU\cite{NIPS2012_4824}, all convolutional kernels are
$3{\times}3$. The fully connected layer is linear (no activation function).
Dropout is set to 0.5.
The network is further normalized using batch normalization \cite{IoffeS15}.
The model has been trained for about 400 epochs using a learning rate of 0.001
and an Adam optimizer with decay rates of 0.9 (beta1) and 0.999 (beta2).

\item The LSTM used in Table \ref{results} is the standard version of \cite{Hochreiter97},
trained for 300 epochs.
For this model all 3 $(x,y)$ coordinate pairs are concatenated to produce
a 6 dimensional feature vector. There are 128 hidden units for a cell, and a fully connected
layer is used to linearly activate the output.

\item For the 2D Spatio-Temporal LSTM we used the variation of \cite{Liu2016}, which is itself a variant of the Multi-dimensional LSTM\cite{Graves2007}. \cite{Liu2016} uses a ``trust-gate'', which filters the input in order to compensate for noise.\\
We apply recurrent dropout as defined in \cite{DBLP:journals/corr/SemeniutaSB16}. It is trained for 150 epochs.
In our model, each cell possesses 64 hidden units and the trust parameter is set to 0.5.
An activation layer takes all cell outputs (from the whole grid) to compute predictions.
\end{itemize}
For every model, we use mini-batches of 64 gestures. We use softmax on the output layers,
and cross-entropy loss.

% \begin{figure}[t]
%    \includegraphics[width=24em]{./img/CNN.png}
%    \caption{\label{CNN} Architecture of the model used. Batch normalization and
%    dropout are applied before max pooling on each convolutional layer.}
% \end{figure}

\noindent
\textbf{Evaluation protocol} ---
We report classification accuracy on the test set, which has been used neither for
training nor for architecture and hyper-parameter optimization. The split between
test data, validation data and training is subject wise. No subject (person) is
in more than one subset of the data.

All optimizations have been optimized using validation error, which is measured with
the leave-one-subject-out (LOSOCV) protocol common in gesture recognition
(test data is \emph{not} used in this protocol).
After optimization of architectures and hyper-parameters, the full
combined training+validation set was used again for retraining the final
model tested on the test set.

% TABLE FOR FINAL SUBMISSION
% \begin{table}[t] \centering
%     \begin{center}
%         \begin{tabular}{cccccc}
%             \arrayrulecolor{cwblue1} \toprule
%             & Methods & Sampling & Data & Accuracy\\
%             &  &  &  augmentation & \\
%             \arrayrulecolor{cwblue1} \toprule
%             A & LSTM \cite{Hochreiter97} & - & X & 58.71\\
%             B & LSTM \cite{Hochreiter97} & Dynamic & X & 73.10\\
%             \arrayrulecolor{cwblue1} \midrule
%             C & 2D-LSTM \cite{Liu2016} & - & X & 60.01\\
%             D & 2D-LSTM \cite{Liu2016} & Dynamic & X & 87.72\\
%             \arrayrulecolor{cwblue1} \midrule
%             E & Convolutional model & - & - &  65.96\\
%             F & Convolutional model & - & X &  73.00\\
%             G & Convolutional model & Uniform & X &  80.95\\
%             H & Convolutional model & Rand. Uniform & X &  80.62\\
%             I & Convolutional model & Dynamic & - &  83.93\\
%             J & Convolutional model & Dynamic & X &  \textbf{89.96}\\
%             \arrayrulecolor{cwblue1} \bottomrule
%         \end{tabular}
%     \end{center}
%     \caption{Results on the proposed multi-touch dataset: different sequential models and ablation study.}
%         \label{results}
% \end{table}
% TABLE FOR ARXIV SUBMISSION (splitted)
\begin{table}[t] \centering
    \begin{center}
        \begin{tabular}{cccccc}
            \arrayrulecolor{cwblue1} \toprule
            & Methods & Dynamic & Data & Accuracy\\
            & & sampling & augmentation &\\
            \arrayrulecolor{cwblue1} \toprule
            A & LSTM \cite{Hochreiter97} & - & X & 58.71\\
            B & LSTM \cite{Hochreiter97} & X & X & 73.10\\
            \arrayrulecolor{cwblue1} \midrule
            C & 2D-LSTM \cite{Liu2016} & - & X & 60.01\\
            D & 2D-LSTM \cite{Liu2016} & X & X & 87.72\\
            \arrayrulecolor{cwblue1} \midrule
            E & Convolutional model & - & - &  65.96\\
            F & Convolutional model & - & X &  73.00\\
            G & Convolutional model & X & - &  83.93\\
            H & Convolutional model & X & X &  \textbf{89.96}\\
            \arrayrulecolor{cwblue1} \bottomrule
        \end{tabular}
    \end{center}
    \caption{Results on the proposed multi-touch dataset: different sequential models and ablation study.}
        \label{results}
\end{table}
\begin{table}[t] \centering
    \begin{center}
        \begin{tabular}{ccc}
            \arrayrulecolor{cwblue1} \toprule
            & Sampling & Accuracy\\
            & type &\\
            \arrayrulecolor{cwblue1} \toprule
            A & No sampling & 73.00\\
            B & Uniform & 80.95\\
            C & Rand. Uniform & 80.62\\
            D & Dynamic & \textbf{89.96}\\
            \arrayrulecolor{cwblue1} \bottomrule
        \end{tabular}
    \end{center}
    \caption{Comparison between different sampling methods for our convolutional model.}
        \label{samplings}
\end{table}

\begin{table}[b]
\begin{center}
\newcommand\items{7}   %Number of classes
\arrayrulecolor{white} %Table line colors
\noindent\begin{tabular}{cc*{\items}{|E}|}
\multicolumn{1}{c}{} &\multicolumn{1}{c}{} &\multicolumn{\items}{c}{Predicted} \\ \hhline{~*\items{|-}|}
\multicolumn{1}{c}{} &\multicolumn{1}{c}{} &\multicolumn{1}{c}{\textbf{1}} &\multicolumn{1}{c}{\textbf{2}} &
\multicolumn{1}{c}{\textbf{3}} &\multicolumn{1}{c}{\textbf{4}} &\multicolumn{1}{c}{\textbf{5}} &\multicolumn{1}{c}{\textbf{6}} &
\multicolumn{1}{c}{\textbf{7}} \\ \hhline{~*\items{|-}|}
\multirow{\items}{*}{\rotatebox{90}{Ground Truth}}
&\textbf{1} & 92 & 6 & 0 & 1 & 0 & 1 & 0 \\ \hhline{~*\items{|-}|}
&\textbf{2} & 7 & 91 & 0 & 0 & 1 & 1 & 0 \\ \hhline{~*\items{|-}|}
&\textbf{3} & 6 & 5 & 64 & 9 & 4 & 11 & 1 \\ \hhline{~*\items{|-}|}
&\textbf{4} & 0 & 1 & 1 & 91 & 6 & 1 & 0 \\ \hhline{~*\items{|-}|}
&\textbf{5} & 0 & 0 & 0 & 2 & 98 & 0 & 0 \\ \hhline{~*\items{|-}|}
&\textbf{6} & 0 & 0 & 0 & 1 & 2 & 97 & 0 \\ \hhline{~*\items{|-}|}
&\textbf{7} & 0 & 1 & 0 & 0 & 0 & 2 & 97 \\ \hhline{~*\items{|-}|}
\end{tabular}
\end{center}
\caption{Confusion matrix on the test set for our dataset. Classes are:
Press Tap, Press Double Tap, Press Scale, Press Twist, Rotation, Scale, Translation}
\label{conf_mat}
\end{table}

\noindent
\textbf{Ablation study} --- In order to assess the effectivness of each part
of the process, we proceed to an incremental evaluation of our method. All the
results are displayed in Table \ref{results}.
\begin{itemize}
\item The two recurrent baselines (LSTM and 2D-LSTM \cite{Liu2016}) perform worse than the convolutional model, which confirms the reasoning that hierarchical respresentations over time are useful for sequences. However, 2D-LSTMs do have several interesting properties, while performing close to the convolutional model (-2 points). They use only 138,631 trainable weights in total (against 446,983 of the CNN), and they can be unrolled on varying
input dimensions. In general, recurrent models are more easily generalizable while CNNs tend to perform better for this task.

\item We trained a model without transition preserving dynamic sampling. To this end, and in order to still have a fixed length representation for the convolutional model, the sequences were cropped or padded to 104 points, which is equivalent to ~1200ms. This value is fitting 95\% of all the dataset gestures.
We observed that longer gestures were most likely held for too long or noisy.

The architecture was optimized for this experiment (again on the validation set),
which resulted in two additional convolutional layers with pooling which are able
to cope with these longer sequences. We added $1{\times}3$ max-pooling over the time dimension.
With 65.96\% on the test set, the model performs much worse then the version with
dynamic pooling. With data augmentation the recognition rate rises to 73.00\%, but is
still far from the 89.96\% we obtain from the full model with dynamic sampling.
\item Data augmentation is an important part of the method, which increases the
invariance of the respresentation with respect to the order of the finger IDs
delivered by the device, as well as the scale of the gestures.
The best performing model w/o augmentation scores at 83.93\%, almost 6 points below
the best performance with augmentation.
\item Our sampling method was also compared to a uniform and a randomized uniform sampling (see Table \ref{samplings}). This last sampling method was defined by uniformly segmenting the sequence, and then picking a sample from each segment using a normal distribution.
\end{itemize}

\noindent
\begin{table}[t]
    \begin{center}
        \begin{tabular}{ccc}
            \arrayrulecolor{cwblue1} \toprule
                     & \multicolumn{2}{c}{Evaluation protocol} \\
            Methods & Number of  & Accuracy\\
            & parameters &\\
            \arrayrulecolor{cwblue1} \toprule
            simple LSTM & 22,663 & 73.10 \\
            4 layers stacked LSTM & 472,839 & 73.65 \\
            Convolutional model & 112,903 & 86.99 \\
            Convolutional model & 446,983 & 89.96 \\
            \arrayrulecolor{cwblue1} \bottomrule
        \end{tabular}
    \end{center}
    \caption{\label{table:num_parameters}Result difference when changing the number of parameters of the models.}
        \label{num_parameters}
\end{table}

\begin{table}[b]
    \begin{center}
        \begin{tabular}{ccc}
            \arrayrulecolor{cwblue1} \toprule
                     & \multicolumn{2}{c}{Evaluation protocol} \\
            Methods & {\footnotesize Leave-one-out} & User-independent \cite{Anthony2012} \\
            \arrayrulecolor{cwblue1} \toprule
            Proposed method & 98.62 & 99.38 \\
            Greedy-5 \cite{Anthony2012} & N/A & 98.0 \\
            \arrayrulecolor{cwblue1} \bottomrule
        \end{tabular}
    \end{center}
    \caption{\label{table:resultsmmg}Results on the MMG Dataset \cite{Anthony2012}.}
        \label{resultsMMG}
\end{table}

\noindent
\textbf{Convolutional and sequential models} --- In order to ensure that the result gap between convolutional and sequential models was not correlated to the number of parameters, we show in Table \ref{num_parameters} the different results obtained with varying number of weights with these two types of models. The weights of the convolutional model were tweaked by changing the number of feature maps, while the sequential model was made ``deeper'' by stacking layers and increasing the hidden state size (more efficient than just increasing the hidden state size).

\noindent
\textbf{Comparison with the state of the art} --- We applied the method on the
MMG dataset, one of the very few standard datasets of this problem.
On this dataset, gestures are complex drawings with a finger or a stylus, performed with a varying number of strokes for a same class. This problem can be seen as symbol recognition. As such, state transitions are not relevant for this task, because temporality does not give meaningful information.
We therefore detected geometric transitions as spatial discontinuities (corners) in the finger trajectories. To this end, we calculated thresholded angles of the spatial gradient and thresholded derivates of these angles. We then sampled 48 points of each gesture using the feature preserving method given in section \ref{sec:sampling} (as opposed to uniform sampling of 96 points done in \cite{Anthony2012}).

For this dataset, we chose an architecture similar to the 6-layers CNN for our own dataset. However, convolutions are 1D, as there is only one stroke at one time on the surface. The architecture is described as follows: 2x(CONV-128+POOL), 2x(CONV-256+POOL), CONV-512, 1x FC. The first CONV layer uses a kernel of size $5$ while the others use a kernel of size$3$. There are a total of 747,536 trainable parameters.

Table \ref{table:resultsmmg} presents the results on this dataset. We used different evaluation protocols: there is no test set for this dataset, so we used the LOSOCV protocol described above for our validation error, as well as the user-independent protocol used in \cite{Anthony2012}. In this protocol, a user among 20 is randomly selected as the test subject, while the training is performed on the 19 other users. The training is done using 9 samples from every class randomly taken from every training user. We then classify one random sample of every class from the test subject. This process is performed 100 times and classification results are averaged.

We can see that the obtained performance of 99.38\% is close to perfect recognition and beats the state of the art of 98.0\% given in \cite{Anthony2012}. This further confirms the interest of our model and sampling procedure.

\textbf{Runtime complexity} --- All computations were done on Nvidia Titan-X Pascal GPUs.
Training the convolutional models on our dataset takes 1h 11min ({$\sim$}400 epochs). Testing a single gesture
takes 1.5 ms, including the dynamic sampling procedure. To assert the portability of this model, runtime on CPU (Intel i7-7700HQ, laptop) is 5ms, only 3.3 times slower than on GPU. This is because input tensors are small, resulting in limited GPU acceleration, I/O being the bottleneck.

%%%%%%%%%%%%%%%%%%%%%%%%%%%%%%%%%%%%%%%%%%%%%%%%%%%%%%%%%%%%%%%%%%%%%%%%%%%%%%%%
\section{Conclusion}

\noindent
We have proposed a novel method for multi-touch gesture recognition based
on learning hierarchical representations from fixed length input. We have
also introduced a dynamic sampling algorithm which preservers sharp features
in the input data. We demonstrated the effectiveness of the method
validating this approach on our dataset of interaction gestures and on an existing
symbolic gesture dataset.

This work is the first step toward a rich and adaptative model for touch surface
interactions. The runtime complexity of the method allows for the development of real-time
applications.

The next challenge after classification is the segmentation of a data flux
in order to recognize multiple gestures at once. This will open our research
to multi-user interactions and long dependency gestures.
In order to process gestures that require an interface update while they are being performed (holding and moving an object in an environment for example), we will focus at some point on early detection processes \cite{Hoai-DelaTorre-IJCV14}.
Another perspective we are looking forward to is reinforcement learning in order
to deal with complex decisions while adapting to model its environment and to
user styles.

% \addtolength{\textheight}{-3cm}   % This command serves to balance the column lengths

{\small
\bibliographystyle{ieee}
\bibliography{refs}

\begin{thebibliography}{10}\itemsep=-1pt

\bibitem{Anthony2012}
L.~Anthony and J.~Wobbrock.
\newblock \$n-protractor: a fast and accurate multistroke recognizer.
\newblock In {\em Proceedings of Graphics Interface 2012}, GI 2012, pages
  117--120, Toronto, Ontario, Canada, 2012. Canadian Human-Computer
  Communications Society.

\bibitem{BaradelArxiv2017}
F.~Baradel, C.~Wolf, and J.~Mille.
\newblock Pose-conditioned spatio-temporal attention for human action
  recognition.
\newblock {\em Pre-print: arxiv:1703.10106}, 2017.

\bibitem{BelongieMalik2002}
S.~Belongie, J.~Malik, and J.~Puzicha.
\newblock Shape matching and object recognition using shape contexts.
\newblock {\em IEEE Transactions on Pattern Analysis and Machine Intelligence},
  24(24):509--521, 2002.

\bibitem{Bengio1994}
Y.~Bengio, P.~Simard, and P.~Frasconi.
\newblock Learning long-term dependencies with gradient descent is difficult.
\newblock {\em Trans. Neur. Netw.}, 5(2):157--166, Mar. 1994.

\bibitem{Chen2014}
Z.~Chen, E.~Anquetil, H.~Mouchere, and C.~Viard-Gaudin.
\newblock {A Graph Modeling Strategy for Multi-touch Gesture Recognition}.
\newblock {\em Proceedings of International Conference on Frontiers in
  Handwriting Recognition, ICFHR}, 2014-December:259--264, 2014.

\bibitem{cho-al-emnlp14}
K.~Cho, B.~van Merri{\"{e}}nboer, {\c C}.~G{\"{u}}l{\c c}ehre, D.~Bahdanau,
  F.~Bougares, H.~Schwenk, and Y.~Bengio.
\newblock Learning phrase representations using rnn encoder--decoder for
  statistical machine translation.
\newblock In {\em Proceedings of the 2014 Conference on Empirical Methods in
  Natural Language Processing (EMNLP)}, pages 1724--1734, Doha, Qatar, Oct.
  2014. Association for Computational Linguistics.

\bibitem{Cirelli2014}
M.~Cirelli and R.~Nakamura.
\newblock A survey on multi-touch gesture recognition and multi-touch
  frameworks.
\newblock In {\em Proceedings of the Ninth ACM International Conference on
  Interactive Tabletops and Surfaces}, ITS '14, pages 35--44, New York, NY,
  USA, 2014. ACM.

\bibitem{tensorflow2015}
M.~A. et~al.
\newblock {TensorFlow}: Large-scale machine learning on heterogeneous systems,
  2015.
\newblock Software available from tensorflow.org.

\bibitem{GehringAGYD17}
J.~Gehring, M.~Auli, D.~Grangier, D.~Yarats, and Y.~N. Dauphin.
\newblock Convolutional sequence to sequence learning.
\newblock {\em CoRR}, abs/1705.03122, 2017.

\bibitem{Graves2007}
A.~Graves, S.~Fern{\'{a}}ndez, and J.~Schmidhuber.
\newblock {Multi-dimensional recurrent neural networks}.
\newblock {\em Artificial Neural Networks--ICANN}, (1):549----558, 2007.

\bibitem{Hoai-DelaTorre-IJCV14}
M.~Hoai and F.~{De la Torre}.
\newblock Max-margin early event detectors.
\newblock {\em International Journal of Computer Vision}, 107(2):191--202,
  2014.

\bibitem{Hochreiter97}
S.~Hochreiter and J.~Schmidhuber.
\newblock Long short-term memory.
\newblock {\em Neural Comput.}, 9(8):1735--1780, Nov. 1997.

\bibitem{IoffeS15}
S.~Ioffe and C.~Szegedy.
\newblock Batch normalization: Accelerating deep network training by reducing
  internal covariate shift.
\newblock {\em CoRR}, abs/1502.03167, 2015.

\bibitem{Kalchbrenner2016}
N.~Kalchbrenner, I.~Danihelka, and A.~Graves.
\newblock {Grid Long Short-Term Memory}.
\newblock In {\em ICLR}, page~14, 2016.

\bibitem{Kin2012}
K.~Kin, B.~Hartmann, T.~DeRose, and M.~Agrawala.
\newblock {Proton++: A Customizable Declarative Multitouch Framework}.
\newblock {\em Uist}, pages 477--486, 2012.

\bibitem{KoutnikGGS14}
J.~Koutn{\'{\i}}k, K.~Greff, F.~J. Gomez, and J.~Schmidhuber.
\newblock A clockwork {RNN}.
\newblock {\em CoRR}, abs/1402.3511, 2014.

\bibitem{NIPS2012_4824}
A.~Krizhevsky, I.~Sutskever, and G.~E. Hinton.
\newblock Imagenet classification with deep convolutional neural networks.
\newblock In F.~Pereira, C.~J.~C. Burges, L.~Bottou, and K.~Q. Weinberger,
  editors, {\em Advances in Neural Information Processing Systems 25}, pages
  1097--1105. Curran Associates, Inc., 2012.

\bibitem{Lafferty2001}
J.~Lafferty, A.~McCallum, and F.~Pereira.
\newblock Conditional random fields: Probabilistic models for segmenting and
  labeling data.
\newblock In {\em International Conference on Machine Learning}, 2001.

\bibitem{Lecun98}
Y.~Lecun, L.~Bottou, Y.~Bengio, and P.~Haffner.
\newblock Gradient-based learning applied to document recognition.
\newblock In {\em Proceedings of the IEEE}, pages 2278--2324, 1998.

\bibitem{LiNeverovaWolfTaylor2017}
F.~Li, N.~Neverova, C.~Wolf, and G.~Taylor.
\newblock {Modout: Learning to Fuse Face and Gesture Modalities with Stochastic
  Regularization}.
\newblock In {\em {FG}}, 2017.

\bibitem{Liu2016}
J.~Liu, A.~Shahroudy, D.~Xu, and G.~Wang.
\newblock {Spatio-Temporal LSTM with Trust Gates for 3D Human Action
  Recognition}.
\newblock 2016.

\bibitem{Lu2012}
H.~L{\"{u}} and Y.~Li.
\newblock {Gesture Coder: A tool for programming multi-touch gestures by
  demonstration}.
\newblock {\em SIGCHI Conference on Human Factors in Computing Systems}, pages
  2875--2884, 2012.

\bibitem{Malik2005}
S.~Malik, A.~Ranjan, and R.~Balakrishnan.
\newblock Interacting with large displays from a distance with vision-tracked
  multi-finger gestural input.
\newblock In {\em Proceedings of the 18th Annual ACM Symposium on User
  Interface Software and Technology}, UIST '05, pages 43--52, New York, NY,
  USA, 2005. ACM.

\bibitem{NeverovaArxiv2016googleIEEEAccess}
N.~Neverova, C.~Wolf, G.~Lacey, L.~Fridman, D.~Chandra, B.~Barbello, G.~Taylor,
  and F.~Nebout.
\newblock Learning human identity from motion patterns.
\newblock {\em {IEEE Access}}, 4:1810--1820, 2016.

\bibitem{NeverovaWolfTaylorNeboutPAMI2016}
N.~Neverova, C.~Wolf, G.~Taylor, and F.~Nebout.
\newblock Moddrop: adaptive multi-modal gesture recognition.
\newblock {\em {IEEE Transactions on Pattern Analysis and Machine
  Intelligence}}, 38(8):1692--1706, 2016.

\bibitem{Rabiner1989}
L.~Rabiner.
\newblock A tutorial on hidden {M}arkov models and selected applications in
  speech recognition.
\newblock {\em Proceedings of the IEEE}, 77(2):257--286, 1989.

\bibitem{Rubine1991}
D.~Rubine.
\newblock Specifying gestures by example.
\newblock In {\em Proceedings of the 18th Annual Conference on Computer
  Graphics and Interactive Techniques}, SIGGRAPH '91, pages 329--337, New York,
  NY, USA, 1991. ACM.

\bibitem{Scholliers2010}
C.~Scholliers, L.~Hoste, B.~Signer, and W.~De~Meuter.
\newblock Midas: A declarative multi-touch interaction framework.
\newblock In {\em Proceedings of the Fifth International Conference on
  Tangible, Embedded, and Embodied Interaction}, TEI '11, pages 49--56, New
  York, NY, USA, 2011. ACM.

\bibitem{DBLP:journals/corr/SemeniutaSB16}
S.~Semeniuta, A.~Severyn, and E.~Barth.
\newblock Recurrent dropout without memory loss.
\newblock {\em CoRR}, abs/1603.05118, 2016.

\bibitem{Wobbrock2007}
J.~O. Wobbrock, A.~D. Wilson, and Y.~Li.
\newblock {Gestures without libraries, toolkits or training: a 1 recognizer for
  user interface prototypes}.
\newblock {\em Proceedings of the 20th annual ACM symposium on User interface
  software and technology UIST 07}, 85(2):159, 2007.

\bibitem{Wu2003}
M.~Wu and R.~Balakrishnan.
\newblock Multi-finger and whole hand gestural interaction techniques for
  multi-user tabletop displays.
\newblock In {\em Proceedings of the 16th Annual ACM Symposium on User
  Interface Software and Technology}, UIST '03, pages 193--202, New York, NY,
  USA, 2003. ACM.

\bibitem{YuHSMM2010}
S.-Z. Yu.
\newblock Hideden semi-markov models.
\newblock {\em Artificial Intelligence}, 174:215--243, 2010.

\bibitem{2013movingpose}
M.~Zanfir, M.~Leordeanu, and C.~Sminchisescu.
\newblock {The Moving Pose: An Efficient 3D Kinematics Descriptor for
  Low-Latency Action Recognition and Detection}.
\newblock In {\em ICCV}, 2013.

\end{thebibliography}
}

%%%%%%%%%%%%%%%%%%%%%%%%%%%%%%%%%%%%%%%%%%%%%%%%%%%%%%%%%%%%%%%%%%%%%%%%%%%%%%%%

\end{document}